\documentclass[11pt]{article}

\usepackage[final]{acl}

\usepackage{times}
\usepackage{latexsym}
\usepackage{booktabs}
\usepackage{multirow}
\usepackage{amsmath}
\usepackage{amssymb}
\usepackage{paralist}
\usepackage{stmaryrd}
\usepackage{booktabs}
\usepackage{paralist}
\usepackage[ruled,vlined]{algorithm2e}
\usepackage{algpseudocode}
\usepackage{todonotes}
\usepackage{xurl}
\usepackage{hyperref} 
\usepackage[T1]{fontenc}
\usepackage{tcolorbox}

\usepackage[utf8]{inputenc}

\usepackage{microtype}

\usepackage{inconsolata}
\usepackage{textcomp}
\usepackage{pifont}
\usepackage{graphicx}
\newcommand{\tool}{\ensuremath{\mathsf{NLICV}}\xspace}
\title{Evaluating LLM Personalization via Semantic Constraint Verification}

\author{
 \textbf{Xuran Li\textsuperscript{1}},
 \textbf{Guanqin Zhang\textsuperscript{1}},
 \textbf{Imran Razzak\textsuperscript{1,2}},
 \\
 \textbf{Hakim Hacid\textsuperscript{3}},
 \textbf{Eleanna Kafeza\textsuperscript{3}},
 \textbf{Hao Xue\textsuperscript{1,4}},
 \textbf{Flora D. Salim\textsuperscript{1}},
\\
 \textsuperscript{1}University of New South Wales,
 \\
 \textsuperscript{2}Mohamed bin Zayed University of Artificial Intelligence,
\\
 \textsuperscript{3}The Technology Innovation Institute,
 \\
 \textsuperscript{4}The Hong Kong University of Science and Technology
}

\begin{document}
\maketitle
\begin{abstract}
Current evaluation paradigms for Large Language Model (LLM) personalization rely heavily on brittle surface-matching metrics or computationally expensive LLM-as-a-judge protocols, both of which lack interpretability. To address these limitations, we introduce \textbf{N}atural \textbf{L}anguage \textbf{I}nference \textbf{C}onstraint \textbf{V}erification (\tool), a scalable, semantically invariant framework that maps sentence meanings to truth-condition sets to verify personalization constraints via {N}atural {L}anguage {I}nference  (NLI) model. Moving beyond binary scoring, \tool categorizes LLM behaviors into four distinct modes: personalization, generalization, sycophancy, and failure. Extensive experiments demonstrate that \tool aligns closely with human annotations while drastically reducing the latency and token costs associated with LLM judges (up to 2100 inference speedup). Finally, through an ablation-based procedure, \tool pinpoints the exact sentences driving the constraint verification, yielding faithful, understandable evidence for its evaluations.

\end{abstract}

\section{Introduction}
Large language models (LLMs) have demonstrated strong general-purpose capabilities across a wide range of tasks \cite{minaee2025largelanguagemodelssurvey, niu2025largelanguagemodelscognitive}. Beyond these general capabilities, personalization has emerged as a critical requirement for improving user satisfaction and enabling context-aware interactions with LLMs. Personalized LLMs are expected to tailor responses according to user preferences, contextual signals, and inferred intents\cite{PersonalizationSurvey}. Many LLMs employ Retrieval-Augmented Generation (RAG) to retrieve user preferences from profiles and incorporate them into the response generation process. However, existing evaluation approaches rely heavily on either human-annotated gold labels or LLM-as-a-Judge protocols \cite{LongLaMP, PersonalLL}, both of which exhibit significant limitations.

\begin{figure}[h!]
    \centering
    \includegraphics[width=1.\linewidth]{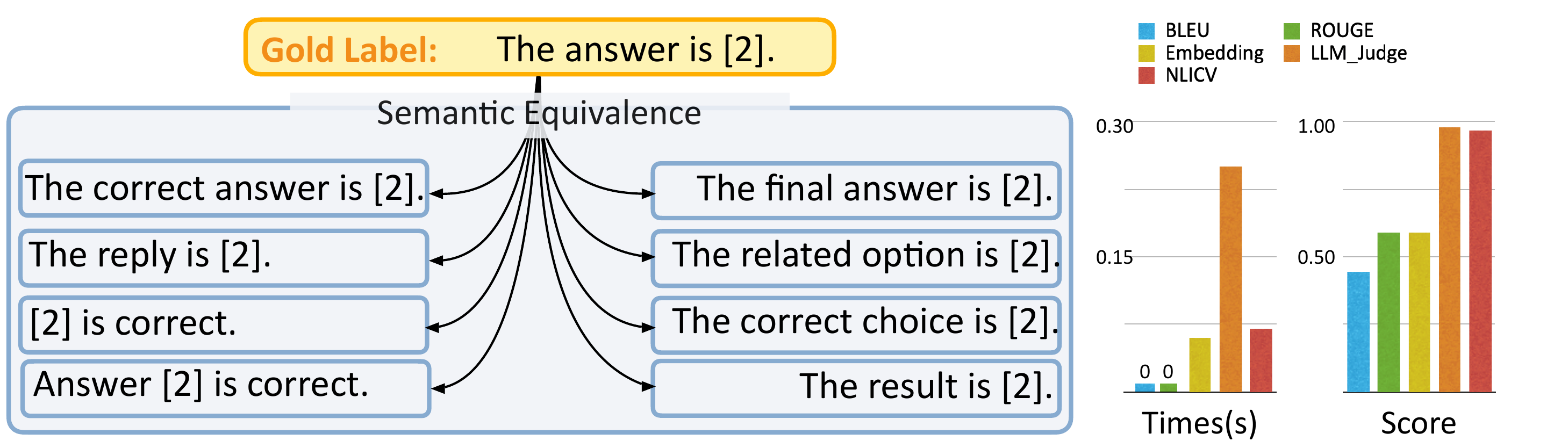}
    \caption{Identical personalized preference evaluation.}
    \label{tab:combined_responses_and_metrics}
\end{figure}

\emph{Limitations of existing works.} 
Current personalization definitions remain under-specified and largely informal, making it difficult to rigorously determine what constitutes a valid personalized response. This lack of formalization propagates directly into evaluation protocols. Annotation-based metrics, including BLEU \cite{papineni-etal-2002-bleu}, ROUGE \cite{dasgupta2024persevalassessingpersonalizationtext}, and embedding similarity, are inherently brittle under semantic invariance. As illustrated in Fig.~\ref{tab:combined_responses_and_metrics}, semantically equivalent responses in the LaMP-1 citation identification task \cite{LaMP} can still receive substantially different scores due to variations in surface realization, leading to inconsistent and unreliable assessments. Although LLM-as-a-Judge approaches partially alleviate this issue, they introduce considerable computational overhead and latency. Moreover, both paradigms offer limited interpretability, providing little insight into why a response is classified as personalized or non-personalized, thereby restricting their usefulness for system analysis and improvement.

\emph{Semantic constraint verification.} 
To address these limitations, we re-examine personalization through the lens of input-output semantic constraints. Specifically, when user context is relevant to a query, a valid personalized response must satisfy two core requirements: (1) accurately addressing the primary query (the $\mathrm{Correct}$ constraint), and (2) faithfully incorporating user-specific preferences (the $\mathrm{Aligned}$ constraint).
Drawing on formal semantics \cite{FormalSemantics}, we represent the meanings of both the user preferences and the model's response as truth-condition sets, i.e., the conditions under which an utterance holds true. The personalization constraints are then formalized through set-theoretic relationships over these truth conditions. In particular, a valid personalized response is required to jointly entail the truth-conditional set of the task answer and that of the user preferences. This reformulation shifts personalization evaluation from subjective matching to a principled semantic verification problem based on structured set relations.

\emph{Natural Language Inference (NLI).} To automatically verify these set-theoretic inclusion relations, we leverage the premise-hypothesis structure of the NLI model. Specifically, the generated LLM response is framed as the premise, while the target semantic constraint serves as the hypothesis. A successful entailment prediction formally establishes that the truth-condition set of the response is a valid subset of that of the constraint. Building on this mechanism, we introduce the \textbf{N}atural \textbf{L}anguage \textbf{I}nference \textbf{C}onstraint \textbf{V}erification (\tool) framework to systematically evaluate LLM personalization via $\mathrm{Correct}$ and $\mathrm{Aligned}$ constraint verification. Moving beyond binary identification, our framework provides a fine-grained taxonomy of LLM behavior by classifying responses into \emph{personalization}, \emph{generalization}, \emph{sycophancy}, and \emph{failure}. Furthermore, to understand why a response satisfies specific constraints, we introduce an ablation-based semantic procedure that pinpoints the critical sentences driving the  verification.

\emph{Empirical Verification.} 
Empirical evaluations confirm that \tool achieves exceptional human alignment and semantic invariance, peaking at $98.00\%$ accuracy on reference-based tasks. Crucially, it eliminates the computational bottleneck of LLM judges by delivering an up to $2,100\times$ inference speedup at zero token cost. While optimizing for large-scale verification, \tool successfully maps responses across four behavioral regimes and utilizes an ablation-based procedure to provide faithful, structural evidences for every constraint check.


\begin{figure*}[h!]
    \centering
    \includegraphics[width=1.\linewidth]{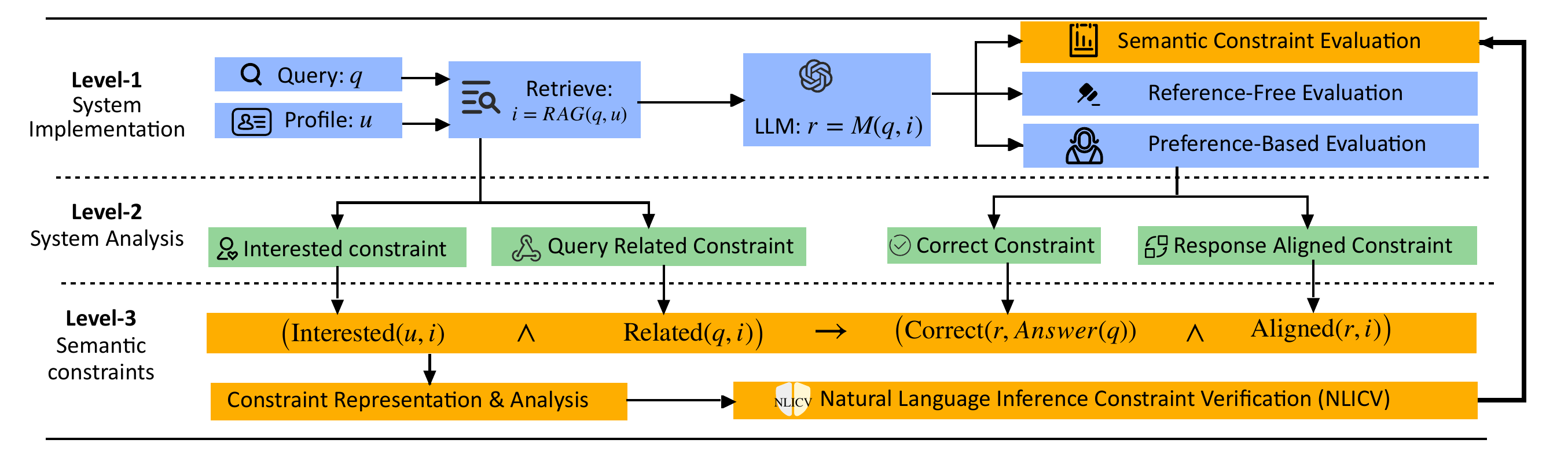}
    \caption{LLM personalization evaluation via semantic constraint verification }
    \label{fig:Referrence-Free Evaluation}
\end{figure*}

\emph{Main Contributions:}
\begin{compactitem}
    \item We propose the \tool framework, establishing a semantic-constraint-driven and theoretically grounded paradigm for evaluating LLM personalization.
    \item We introduce a fine-grained taxonomy of LLM behavior. \tool systematically classifies responses into \emph{personalization}, \emph{generalization}, \emph{sycophancy}, and \emph{failure}, by jointly verifying $\mathrm{Correct}$ and $\mathrm{Aligned}$ constraints.
    \item We develop an ablation-based semantic procedure that ensures evaluation transparency. This method pinpoints the critical sentences driving constraint verification, explicitly explaining precisely why a response satisfies the specific constraint.
    \item Extensive evaluations validate \tool's high human alignment and semantic invariance under perturbations. Notably, the framework bypasses the computational limits of LLM Judges, achieving an up to $2,100\times$ inference speedup at 0 token cost while remaining highly effective for large-scale verification.
\end{compactitem}

\section{Proposed framework: \tool}

This section details the \tool framework across three core stages: formalizing the semantic constraints of LLM personalization, establishing a rigorous set-theoretic verification method, and practically operationalizing these constraints via NLI.


\subsection{Semantic Constraints}

Personalization aims to adapt LLM outputs to user-specific information. Let $q \in \mathcal{Q}$ denote a user query, $u \in \mathcal{U}$ a user profile, and $r \in \mathcal{R}$ a generated response. A personalized language model can be formulated as $r = M(q, i)$, where $i = \mathrm{RAG}(q, u)$.
Here, $M$ denotes the LLM and $\mathrm{RAG}$ denotes a retrieval function that extracts user-specific preference information $i$ relevant to query $q$ from the user profile $u$, as illustrated in Level 1 of Fig.~\ref{fig:Referrence-Free Evaluation}. 

To characterize the personalization process, \tool defines four semantic constraints: $\mathrm{Interested}(u,i), \mathrm{Related}(q,i), \mathrm{Correct}(r,a)$,and $\mathrm{Aligned}(r,i)$.
The first two constraints govern the input stage, while the latter two govern the output stage. 
The overall process can be expressed:

\begin{equation}
\label{eq:personalized process}
\begin{aligned}
    &(\mathrm{Interested}(u,i) \land \mathrm{Related}(q,i)) \rightarrow \\
    &\quad (\mathrm{Correct}(r,a) \land \mathrm{Aligned}(r,i)).
\end{aligned}
\end{equation}
In Eq.~\ref{eq:personalized process}, $\mathrm{Interested}(u,i)$ and $\mathrm{Related}(q,i)$ constrain the retrieval process by ensuring that the retrieved information $i$ originates from the user profile $u$ and remains relevant to the query $q$. While, $\mathrm{Correct}(r,a)$ and $\mathrm{Aligned}(r,i)$ characterize the validity of the generated response: $\mathrm{Correct}(r,a)$ measures whether $r$ is semantically consistent with the underlying answer space $a$, while $\mathrm{Aligned}(r,i)$ measures the extent to which $r$ successfully incorporates user-specific information $i$. To formalize correctness independently of any specific reference response, we introduce a latent semantic answer variable, $a = \mathrm{Answer}(q)$, which represents the set of all valid answers to query $q$. 
\begin{figure}[h!]
    \centering
    \includegraphics[width=0.92\linewidth]{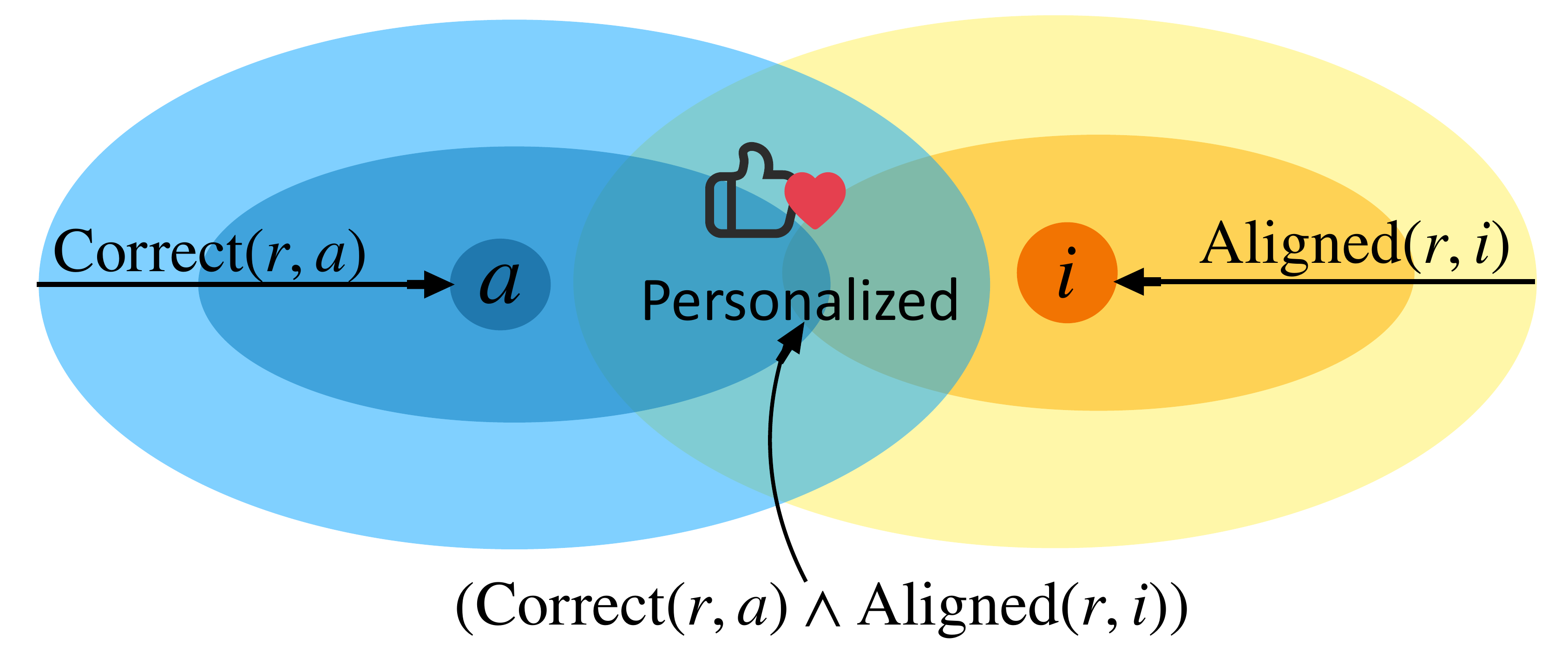}
    \caption{Personalized response space}
    \label{fig:optimization}
\end{figure}

Since this work focuses primarily on output evaluation, we assume that a valid personalized response should maximize user alignment while remaining strictly consistent with the underlying answer space, illustrated in Eq,~\ref{eq:reference_label} and Fig.~\ref{fig:optimization}: 
\begin{equation}
\label{eq:reference_label}
\begin{aligned}
    r^* &= \arg\max_{r \in \mathcal{R}} \mathrm{Aligned}(r,i) \\
    &\quad \text{s.t.} \quad \mathrm{Correct}(r,a) \ge \tau_{\mathrm{correct}},
\end{aligned}
\end{equation}
where $\tau_{\mathrm{correct}}$ denotes a strict correctness threshold. 

Eq.~\ref{eq:reference_label} explicitly characterizes personalized responses in terms of semantic constraint satisfaction rather than surface-form similarity.
Once $r^*$ is formally defined through Eq.~\ref{eq:reference_label}, direct comparison against a gold label becomes unnecessary. Instead, a generated response $r$ is considered valid if it satisfies Eq.~\ref{eq:reference_label}. Consequently, the evaluation paradigm shifts from surface-level similarity estimation to semantic constraint verification. 

\subsection{Constraint Verification}

To verify these semantic constraints, \tool adopts a two-level approach grounded in formal semantics \cite{FormalSemantics}. First, truth-conditional semantics provides a rigorous mathematical representation of natural language meaning. Second, semantic entailment relations derived from set theory enable semantic constraint verification.

Let $W$ denote the set of possible worlds, where each world $w \in W$ represents a logically coherent state of affairs. In formal semantics, the meaning of a sentence is characterized by its truth conditions, the specific set of worlds where the sentence is true.
We adopted the standard type-theoretic notation, and let $t=\{1,0\}$ and $s$ denote truth values and the type of possible worlds. A proposition, therefore, has type $\langle s,t\rangle$, corresponding to a function mapping worlds to truth values. For a sentence $S$, its semantic interpretation is defined as:
$\llbracket S \rrbracket : W \rightarrow \{1,0\}$,
where $\llbracket S \rrbracket(w)=1$ if and only if $S$ is true in world $w$.
Equivalently, a sentence can be represented by its \emph{truth-condition set}:
$\mathcal{T}(S)=\{\, w \in W \mid \llbracket S \rrbracket(w)=1 \,\},$
of which the semantic meaning is strictly identified with the set of worlds aligned with the content of sentences

Truth-condition sets naturally induce logical relations between natural language expressions,
such as entailment:$S_1 \models S_2 \iff \mathcal{T}(S_1)\subseteq \mathcal{T}(S_2)$.  
This set-theoretic relation provides the formal foundation for constraint verification. Let $\llbracket r \rrbracket$, $\llbracket a \rrbracket$, and $\llbracket i \rrbracket$ denote the truth-condition sets of the generated response $r$, the latent answer space $a$, and the retrieved user-specific information $i$. The semantic personalization constraints are defined as follows: 
\begin{align}
    \mathrm{Correct}(r,a) &\iff \llbracket r \rrbracket \subseteq \llbracket a \rrbracket, \label{eq:correct} \\
    \mathrm{Aligned}(r,i) &\iff \llbracket r \rrbracket \subseteq \llbracket i \rrbracket. \label{eq:aligned}
\end{align}
A valid personalized response must entail both the factual answer (Eq.~\ref{eq:correct}) and the user preference (Eq.~\ref{eq:aligned}). Eq.~\ref{eq:correct} ensures that the generated response preserves the semantic truth conditions of the valid answer space (i.e., every world where the response is true is also a world where the correct answer holds). Eq.~\ref{eq:aligned} requires that the response's truth conditions be strictly encompassed by the user preference set. The response is semantically constrained to exist within the intersection of the answer space and the preference space: $\llbracket r \rrbracket \subseteq (\llbracket a \rrbracket \cap \llbracket i \rrbracket)$.

\textbf{Illustrative Example.}
Tab.~\ref{tab:combined_responses_and_metrics} presents eight semantically equivalent responses. Consider the gold label: $S_g :=$ \emph{`The answer is [2].'} and the response $S_r :=$ \emph{`[2] is correct.'}
Their lexical realizations differ, but both evaluate to true in exactly the same worlds: the worlds in which option [2] represents the correct answer. Formally, their truth-condition sets are identical:$\mathcal{T}(S_g) = \mathcal{T}(S_r)$.
Because $S_g \equiv S_r$, the response flawlessly preserves the semantic truth conditions of the latent answer space $a$. Therefore, it satisfies Eq.\ref{eq:correct}.

Now suppose the retrieved user preference information is $i :=$ \emph{`Hardware Characterization'}, and the response $r$ elaborates on option [2] as follows: \emph{`A low-cost wear-leveling algorithm for block-mapping solid-state disks.'} The semantic content $r$ logically entails the hardware-related preference space represented by $\llbracket i \rrbracket$, because domain concepts such as wear-leveling, block-mapping, and solid-state disks are inherent subsets of hardware characterization. In possible-world semantics, every world in which this response is true is necessarily a world in which hardware characterization is discussed. Hence, 
it satisfies Eq.~\ref{eq:aligned}

In contrast, a response emphasizing unrelated software applications or abstract theoretical discussions may still satisfy the correctness constraint (by identifying option [2]), while failing the alignment constraint. Its semantic content would not entail the user preference space ($\llbracket r \rrbracket \not\subseteq \llbracket i \rrbracket$). Consequently, these constraints decouple personalization evaluation into two independent, verifiable dimensions: correctness guarantees semantic consistency with the valid answer space, while alignment guarantees logical inclusion within the user preferences.

\subsection{Constraint Operationalization}

While possible-world semantics provides a precise theoretical foundation, the direct mathematical manipulation of truth-condition sets is computationally intractable. To operationalize this framework, \tool bridges set-theoretic relations via textual entailment, executing the verification through Natural Language Inference (NLI) models.

In NLI, a premise $P$ and a hypothesis $H$ are evaluated to determine whether $P$ logically entails $H$ ( $P \models H$). Semantic entailment serves as the direct computational counterpart to set inclusion:
$P \models H \iff \mathcal{T}(P) \subseteq \mathcal{T}(H)$.

By reducing each semantic constraint to a $(P,H)$ verification task, \tool efficiently computes LLM personalization evaluation at scale. The strict subset relations are instantiated via textual entailment:

\emph{1) Factual Correctness:} $\mathrm{Correct}(r,a) \iff (r \models a)$. The response $r$ should logically entail the core answer $a$. This ensures factual consistency and naturally accommodates personalization.  The addition of user-specific details does not violate the entailment of the core facts.

\emph{2) Personalized Alignment:} $\mathrm{Aligned}(r,i) \iff (r \models i)$.The alignment is defined as strict logical inclusion ($\llbracket r \rrbracket \subseteq \llbracket i \rrbracket$). This guarantees that the semantic content of the generated text is entirely encompassed by the target preference space.



These entailment relations are evaluated via the NLI model. For each $(P,H)$ pair, a semantic constraint is satisfied if the predicted entailment probability, $S_{\mathrm{ent}}(P,H) = \Pr(P \models H)$, meets a predefined confidence threshold:$S_{\mathrm{ent}}(P,H) \ge \tau_{\mathrm{sem}}.$


\tool provides a principled bridge between formal semantic theory and practical LLM evaluation. The possible-world framework mathematically defines the exact meaning of correctness and alignment, while NLI offers a scalable, computable execution of those set relations. \tool transforms the personalization evaluation to the constraint verification problem grounded in semantic reasoning.

\section{Personalization Confusion Matrix}

\begin{figure}[h]
    \centering
    \includegraphics[width=0.9\linewidth]{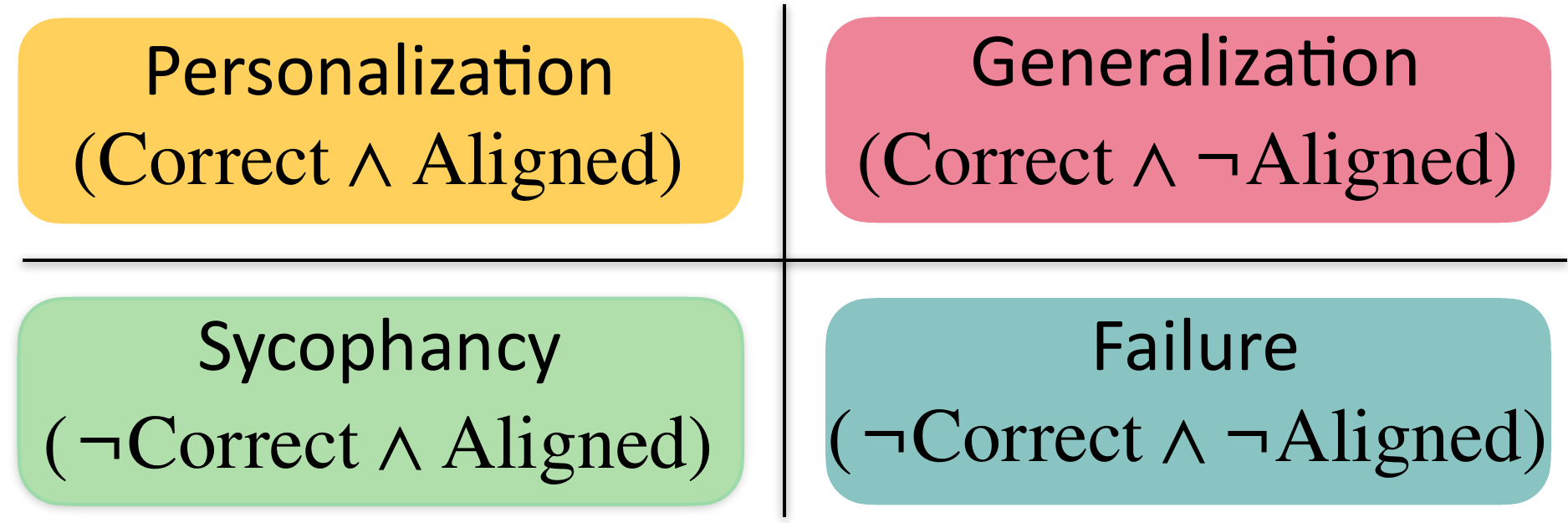}
    \caption{Personalization confusion matrix}
    \label{fig:Personalization Confusion Matrix}
\end{figure}

Moving beyond binary evaluation, \tool characterizes LLM personalization with high granularity by jointly verifying $\mathrm{Correct}$ and $\mathrm{Aligned}$ constraints. The personalization confusion matrix visualizes this in (Fig.~\ref{fig:Personalization Confusion Matrix}), which maps model responses into four semantically distinct behaviors.



\begin{compactitem}
    \item \textbf{Personalization}($\mathrm{Correct} \land \mathrm{Aligned}$):  
    The response is both factually correct within the valid answer space and consistent with the user-specific preference constraints.  This represents the desired personalized behavior.

    \item \textbf{Generalization} ($\mathrm{Correct} \land \neg\mathrm{Aligned}$):  
    The response is factually correct but fails to incorporate user preferences. 
    The LLM behaves as a generic, non-personalized assistant.

    \item \textbf{Sycophancy} ($\neg\mathrm{Correct} \land \mathrm{Aligned}$):  
    The response conforms to the user's preferences while sacrificing factual correctness. 
    This captures the well-known failure mode in which an LLM prioritizes agreement with the user over objective validity.

    \item \textbf{Failure} ($\neg\mathrm{Correct} \land \neg\mathrm{Aligned}$):  
    The response is neither factually correct nor aligned with the user’s preferences.
    representing a complete failure of generation.
\end{compactitem}

This taxonomy extends evaluation beyond a single scalar metric and enables a more nuanced analysis of LLM behavior, disentangling faithful personalization from undesirable phenomena such as over-generalization and sycophancy.

\subsection{Inference Evidences}
To provide structural inference evidences, \tool introduces an ablation-based semantic interaction analysis. This procedure identifies the minimal semantic units driving the $(P,H)$ entailment, and exposes the underlying evidence-constraint dependency structure that governs the verification.

We first decompose the premise $P$ and hypothesis $H$ into sets of atomic propositions, denoted as $P = \{p_\theta\}_{\theta=1}^{m}$ and $H = \{h_\phi\}_{\phi=1}^{n}$. Here, an atomic proposition is formally defined as the smallest semantically indivisible unit capable of expressing a complete and verifiable truth condition. In truth-conditional semantics, a text's meaning equates to the set of possible worlds where the text is true. Because the premise and hypothesis act as logical conjunctions of their atomic propositions, their truth-conditional sets are expressed as:
$\llbracket P \rrbracket = \bigcap_{\theta=1}^{m} \llbracket p_\theta \rrbracket, \text{ and }\llbracket H \rrbracket = \bigcap_{\phi=1}^{n} \llbracket h_\phi \rrbracket. $

This reflects semantic compositionality: adding a proposition imposes a stricter constraint, monotonically shrinking the set of admissible worlds. Conversely, removing a proposition relaxes the semantic boundaries, expanding the compatible world set. This inherent monotonicity provides the theoretical grounding for our ablation procedure.

A two-stage ablation procedure is applied to map the structural dependencies between premise evidence and hypothesis constraints. By observing entailment shifts under controlled ablation, \tool produces logic-grounded explanations for the final constraint verification.
The first-stage ablation evaluates the necessity of individual semantic units on both the premise and hypothesis sides. 

\emph{1). Premise Ablation (Evidence Necessity).}
For each premise proposition $p_\theta$,  construct an ablated premise: $P_{-\theta} = P \setminus \{p_\theta\}.$
Removing a conjunct weakens the premise and expands the world set: $\llbracket P \rrbracket \subseteq \llbracket P_{-\theta} \rrbracket,$ then test whether the expanded world set continues to entail the hypothesis.
\begin{compactitem}
    \item {Critical Premise Unit:}  
    A unit $p_\theta$ is {critical} if removing it breaks entailment: $\llbracket P_{-\theta} \rrbracket \not\subseteq \llbracket H \rrbracket.$ such a unit constitutes indispensable semantic evidence for satisfying the target constraint.
    \item {Non-Critical Premise Unit.}  
    A unit $p_\theta$ is {non-critical} if entailment remains valid after removal: $\llbracket P_{-\theta} \rrbracket \subseteq \llbracket H \rrbracket.$ These units are semantically redundant to the entailment relation.
\end{compactitem}
\emph{2). Hypothesis Ablation (Constraint Sensitivity).}
Similarly, for each hypothesis proposition $h_\phi$, construct an ablated hypothesis: $H_{-\phi} = H \setminus \{h_\phi\}$, and test whether this relaxation weakens its original semantic boundary.

\begin{compactitem}
    \item {Critical Hypothesis Unit.}  $h_\phi$ is {critical} if removing it breaks entailment: $\llbracket H_{-\phi}\rrbracket \not\subseteq \llbracket H \rrbracket.$ 
    \item {Non-Critical Hypothesis Unit.} $h_\phi$ is {non-critical} if it is logically entailed by the remaining hypothesis units: $\llbracket H_{-\phi} \rrbracket \subseteq \llbracket H \rrbracket.$ 
\end{compactitem}

While it identifies individually necessary semantic units, it does not reveal the mappings between specific premise propositions and hypothesis constraints. To localize these dependencies, \tool performs second-order ablation over jointly ablated configurations of critical units, testing whether entailment is restored: $\llbracket P_{-\theta} \rrbracket \subseteq \llbracket H_{-\phi} \rrbracket$.

\emph{Evidence-Constraint Pair.}
A pair $(p_\theta, h_\phi)$ is formally defined as an \emph{evidence-constraint pair} if:
$ \llbracket P_{-\theta} \rrbracket \not\subseteq \llbracket H \rrbracket\ \text{and}\  \llbracket P_{-\theta} \rrbracket \subseteq \llbracket H_{-\phi} \rrbracket$.

The first condition confirms that $p_\theta$ is indispensable under the hypothesis. The second condition proves that ablating $h_\phi$ neutralizes the entailment violation caused by the removal of $p_\theta$. Therefore, the pair $(p_\theta, h_\phi)$ encodes a direct semantic dependency, mapping a proposition $p_\theta$ as the exact evidence satisfying constraint $h_\phi$.

This two-stage ablation procedure produces fine-grained and understandable explanations of semantic verification and explicitly identifies the responsible evidence for satisfying constraints. We defer further algorithmic details and comprehensive case studies to Appendix \ref{Ablation_Algorithm} and Experiment \ref{Case_Study}.

\begin{table*}[h]
\centering
\caption{Evaluation of human alignment and semantic invariance under semantic-preserving lexical perturbations.}
\label{tab:alignment_metrics}

\resizebox{\textwidth}{!}{%
\begin{tabular}{l ccc ccc ccc ccc ccc ccc ccc ccc}
\toprule

\multirow{2}{*}{\textbf{Rate}}
& \multicolumn{3}{c}{\textbf{BLEU}}
& \multicolumn{3}{c}{\textbf{ROUGE}}
& \multicolumn{3}{c}{\textbf{Embedding}}
& \multicolumn{3}{c}{\textbf{DeBERTa-Large}}
& \multicolumn{3}{c}{\textbf{DeBERTa-Small}}
& \multicolumn{3}{c}{\textbf{MiniLM2}}
& \multicolumn{3}{c}{\textbf{DistilRoBERTa}}
& \multicolumn{3}{c}{\textbf{RoBERTa}} \\

\cmidrule(lr){2-4}
\cmidrule(lr){5-7}
\cmidrule(lr){8-10}
\cmidrule(lr){11-13}
\cmidrule(lr){14-16}
\cmidrule(lr){17-19}
\cmidrule(lr){20-22}
\cmidrule(lr){23-25}

& \textbf{Pos$\uparrow$}
& \textbf{Neg$\downarrow$}
& \textbf{Acc.}

& \textbf{Pos$\uparrow$}
& \textbf{Neg$\downarrow$}
& \textbf{Acc.}

& \textbf{Pos$\uparrow$}
& \textbf{Neg$\downarrow$}
& \textbf{Acc.}

& \textbf{Pos$\uparrow$}
& \textbf{Neg$\downarrow$}
& \textbf{Acc.}

& \textbf{Pos$\uparrow$}
& \textbf{Neg$\downarrow$}
& \textbf{Acc.}

& \textbf{Pos$\uparrow$}
& \textbf{Neg$\downarrow$}
& \textbf{Acc.}

& \textbf{Pos$\uparrow$}
& \textbf{Neg$\downarrow$}
& \textbf{Acc.}

& \textbf{Pos$\uparrow$}
& \textbf{Neg$\downarrow$}
& \textbf{Acc.}
\\

\midrule

\textbf{0.00}
& 1.00 & 0.83 & 50.00
& 1.00 & 0.81 & 50.00
& 1.00 & 0.80 & 50.00
& 0.99 & 0.49 & \textbf{99.67}
& 0.94 & 0.44 & 96.83
& 0.70 & 0.20 & 68.50
& 0.69 & 0.21 & 68.50
& 0.69 & 0.18 & 69.00
\\

\textbf{0.20}
& 0.71 & 0.58 & 57.17
& 0.80 & 0.63 & 59.67
& 0.84 & 0.65 & 59.83
& 0.98 & 0.49 & \textbf{98.17}
& 0.93 & 0.44 & 95.33
& 0.68 & 0.21 & 66.33
& 0.70 & 0.22 & 70.67
& 0.70 & 0.22 & 71.17
\\

\textbf{0.40}
& 0.46 & 0.36 & 61.00
& 0.60 & 0.44 & 72.00
& 0.66 & 0.51 & 71.00
& 0.98 & 0.49 & \textbf{98.50}
& 0.93 & 0.44 & 96.50
& 0.67 & 0.21 & 66.17
& 0.67 & 0.23 & 68.33
& 0.69 & 0.24 & 70.33
\\

\textbf{0.60}
& 0.37 & 0.25 & 52.83
& 0.51 & 0.34 & 58.17
& 0.58 & 0.41 & 72.17
& 0.97 & 0.50 & \textbf{96.67}
& 0.93 & 0.45 & 95.33
& 0.67 & 0.21 & 68.33
& 0.66 & 0.23 & 67.50
& 0.69 & 0.25 & 71.00
\\

\textbf{0.80}
& 0.35 & 0.25 & 51.50
& 0.50 & 0.33 & 57.83
& 0.57 & 0.40 & 71.50
& 0.97 & 0.50 & \textbf{97.00}
& 0.94 & 0.45 & 95.17
& 0.67 & 0.21 & 67.67
& 0.66 & 0.23 & 67.33
& 0.69 & 0.26 & 70.17
\\

\textbf{1.00}
& 0.35 & 0.25 & 50.83
& 0.50 & 0.33 & 58.00
& 0.57 & 0.41 & 70.00
& 0.97 & 0.50 & \textbf{98.00}
& 0.94 & 0.44 & 95.50
& 0.66 & 0.21 & 67.00
& 0.65 & 0.23 & 66.50
& 0.68 & 0.26 & 69.00
\\

\midrule

\textbf{Std}
& 0.26 & 0.24 & 4.30
& 0.21 & 0.20 & 7.11
& 0.18 & 0.16 & 8.98
& \textbf{0.01} & \textbf{0.01} & \textbf{1.08}
& 0.01 & 0.01 & 0.70
& 0.01 & 0.00 & 0.99
& 0.02 & 0.01 & 1.44
& 0.01 & 0.03 & 0.94
\\

\textbf{Mean}
& 0.54 & 0.42 & 53.89
& 0.65 & 0.48 & 59.28
& 0.70 & 0.53 & 65.75
& 0.98 & 0.49 & \textbf{98.00}
& 0.94 & 0.44 & 95.78
& 0.68 & 0.21 & 67.33
& 0.67 & 0.22 & 68.14
& 0.69 & 0.24 & 70.11
\\

\bottomrule
\end{tabular}%
}
\end{table*}

\section{Experimental Evaluation}
Our evaluation  is structured to address three core objectives: (i) quantifying the alignment of \tool and traditional reference-based metrics with human judgments under semantic-preserving perturbations; (ii) benchmarking \tool against LLM-as-a-Judge baselines in a strictly gold-label-free setting; and (iii) demonstrating the inference evidences of \tool through the ablation-based algorithm.

\subsection{Evaluation Settings}
\emph{Datasets.} We assess human alignment using the Citation Identification, Movie Tagging, and Product Rating tasks from the LaMP benchmark \cite{LaMP}, and leverage the Massive Multitask Language Understanding (MMLU) benchmark \cite{hendryckstest2021} to compare \tool against LLM judges in evaluating personalization via semantic constraints.

\emph{Models \& Implementation.} All experiments are conducted on NVIDIA H100 NVL GPUs (96 GB HBM3).  The \tool framework is implemented in Python, setting the semantic confidence threshold to $\tau_{\mathrm{sem}}=0.5$. As the core NLI filter, we evaluate DeBERTa-v3 (Large and Small) \cite{NLI_model}, MiniLMv2 \cite{wang2021minilmv2multiheadselfattentionrelation}, DistilRoBERTa \cite{Sanh2019DistilBERTAD}, and RoBERTa \cite{liu2019robertarobustlyoptimizedbert}. For the LLM-as-a-Judge baselines, we utilize Qwen-3-8B \cite{qwen3technicalreport}, Llama-3-8B \cite{llama3modelcard}, DeepSeek-R1 \cite{deepseekai2025deepseekr1incentivizingreasoningcapability}, and Falcon-H1R \cite{falcon-h1r} to classify responses according to our Personalization Confusion Matrix. For reproducibility and anonymous review, all code, prompts, and evaluation scripts are released in an anonymous repository~\cite{nlicv_anonymous_repo}


\subsection{Human Alignment Evaluation under Semantic Invariance Perturbations}
To assess human alignment and semantic robustness, we frame personalization evaluation as a binary verification task. Within this setup, gold-label personalized responses from the LaMP benchmark act as positive samples, whereas their non-personalized equivalents serve as negative samples. An optimal metric must demonstrate high discriminative power, consistently assigning high scores to personalized text while effectively penalizing non-personalized outputs.

We benchmark the proposed \tool framework against established reference-based metrics, namely BLEU~\cite{papineni-etal-2002-bleu}, ROUGE~\cite{dasgupta2024persevalassessingpersonalizationtext}, and embedding similarity. To rigorously measure semantic invariance, we introduce semantic-preserving lexical perturbations via synonym replacement across varying rephrase rates (from 0.0 to 1.0). A robust evaluation framework must yield consistent judgments regardless of surface-level modifications. \textbf{Pos} denotes the average score assigned to positive (personalized) responses, while \textbf{Neg} denotes the average score assigned to negative (non-personalized) responses.



The Tab.~\ref{tab:alignment_metrics} reveals the fundamental inability of traditional metrics to capture semantic consistency. At a zero rephrase rate, BLEU, ROUGE, and embedding similarity exhibit poor discriminative capacity, assigning spuriously high scores to lexically similar negative labels (e.g., $0.83$ average BLEU). As lexical perturbation increases, these metrics fail to recognize synonym-replaced semantic equivalences, aggressively penalizing positive instances. Across all rephrase rates, baseline accuracy stagnates near random-guess performance ($\sim 50\%$), proving that surface and embedding similarities are inadequate for robust verification.

\tool entirely bypasses this limitation. It evaluates structural logic rather than lexical overlap and maintains strict semantic invariance. The DeBERTa-Large variant achieves a peak mean accuracy of $98.00\%$ (std: $1.08\%$), maintaining nearly identical performance across all rephrase rates.

These findings strongly corroborate our core hypothesis: effective personalization evaluation requires semantic constraint verification, not surface-level matching. Through explicit entailment modeling, \tool significantly outperforms traditional reference-based metrics in robustness, stability, and alignment with human judgment.

\subsection{Semantic Constraint Evaluation}

\begin{table}[h!]
\centering
\caption{Correctness ($S_{\text{c}}$) and Alignment ($S_{\text{a}}$) constraint scores across the four behavioral regimes of the personalization confusion matrix.}
\resizebox{0.9\linewidth}{!}{%
\begin{tabular}{l cc cc cc cc}
\toprule
& \multicolumn{2}{c}{\textbf{Generalized}} & \multicolumn{2}{c}{\textbf{Personalized}} & \multicolumn{2}{c}{\textbf{Sycophancy}} & \multicolumn{2}{c}{\textbf{Failure}} \\
\cmidrule(lr){2-3} \cmidrule(lr){4-5} \cmidrule(lr){6-7} \cmidrule(lr){8-9}
\textbf{Model} & $S_{\text{c}}$ & $S_{\text{a}}$ & $S_{\text{c}}$ & $S_{\text{a}}$ & $S_{\text{c}}$ & $S_{\text{a}}$ & $S_{\text{c}}$ & $S_{\text{a}}$ \\
\midrule
Qwen3-8B         & 0.85 & 0.62 & 0.39 & 0.44 & 0.28 & 0.32 & 0.34 & 0.33 \\ 
DeepSeek-R1      & 0.80 & 0.80 & 0.47 & 0.48 & 0.31 & 0.32 & 0.31 & 0.33 \\
Llama-3-8B       & 0.77 & 0.77 & 0.50 & 0.50 & 0.42 & 0.42 & 0.44 & 0.44 \\ 
Falcon-H1R       & 0.72 & 0.72 & 0.42 & 0.48 & 0.23 & 0.27 & 0.26 & 0.25 \\ 
\midrule
\textbf{DeBERTa-Large} & \textbf{0.96} & \textbf{0.03} & \textbf{0.58} & \textbf{0.68} & \textbf{0.18} & \textbf{0.68} & \textbf{0.18} & \textbf{0.03} \\
DeBERTa-Small    & 0.94 & 0.05 & 0.49 & 0.57 & 0.12 & 0.57 & 0.23 & 0.05 \\
MiniLM2          & 0.79 & 0.03 & 0.11 & 0.13 & 0.07 & 0.13 & 0.17 & 0.03 \\
DistilRoBERTa    & 0.85 & 0.03 & 0.16 & 0.16 & 0.10 & 0.17 & 0.23 & 0.03 \\
RoBERTa          & 0.95 & 0.04 & 0.38 & 0.36 & 0.13 & 0.36 & 0.19 & 0.03 \\
\bottomrule
\end{tabular}%
}
\label{tab:model_scores}
\end{table}

To assess personalization verification under semantic constraints, we design a preference-free evaluation on the MMLU benchmark. By clustering query topics, we construct 10 distinct topic distributions to simulate 10 users with unique domain preferences, sampling 10\% of the MMLU test set as our query corpus. To populate the \emph{Personalization Confusion Matrix}, we instruct Qwen3-8B to generate four distinct response variants per query using the ground truth answers and distractors.

\emph{1). Personalized:} Correctly addresses the query while explicitly aligning with user preferences.
\emph{2). Generalized:} Correctly addresses the query but omits personalized information.
\emph{3). Sycophancy:} Provides an incorrect answer while explicitly appealing to user preferences.
\emph{4). Failure:} Provides an incorrect answer devoid of personalization.

In this environment, both LLM-as-a-judge baselines and \tool are tasked with categorizing each response by checking the correctness and alignment constraints. We evaluate their performance based on \emph{classification accuracy} and \emph{computational efficiency}, and standardize the evaluation budget to one hour per user profile ($\sim$10 hours total).  We systematically record inference latency and token consumption. The detailed instruction templates are attached in Appendix \ref{Templates}.

\begin{table*}[!h]
\centering
\caption{
Classification accuracy of \tool versus LLM-as-a-Judge baselines for reference-free personalization evaluation, with a time budget of one hour per user profile ($\sim$10 hours in total).}
\label{tab:performance_results}
\resizebox{0.9\textwidth}{!}{
\begin{tabular}{@{} l r rrrr rrrr r rrrr @{}}
\toprule

\multirow{2}{*}{\textbf{Model}} & 
\multirow{2}{*}{\textbf{Items}} & 
\multicolumn{2}{c}{\textbf{Generalized}} & 
\multicolumn{2}{c}{\textbf{Personalized}} & 
\multicolumn{2}{c}{\textbf{Sycophancy}} & 
\multicolumn{2}{c}{\textbf{Failure}} & 
\textbf{Mean} & 
\multicolumn{2}{c}{\textbf{Time (s)}} & 
\multicolumn{2}{c}{\textbf{Tokens}} \\

\cmidrule(lr){3-4} \cmidrule(lr){5-6} \cmidrule(lr){7-8} \cmidrule(lr){9-10} \cmidrule(lr){12-13} \cmidrule(l){14-15}

& & \textbf{Score} & \textbf{Acc.} & \textbf{Score} & \textbf{Acc.} & \textbf{Score} & \textbf{Acc.} & \textbf{Score} & \textbf{Acc.} & \textbf{Acc.} & \textbf{Avg.} & \textbf{Total} & \textbf{Avg.} & \textbf{Total} \\ \midrule

\textbf{Qwen3-8B}         &  1,723 & 0.74 & 18.69 & 0.42 & 33.08 & 0.30 &  6.79 & 0.34 & 65.76 & 31.08 &  20.97 & 36,134.73 & 4,763.17 & 8,206,940 \\
\textbf{DeepSeek-R1}      &    627 & \textbf{0.80} &  0.00 & 0.48 & 46.25 & 0.32 &  0.48 & 0.32 & 75.92 & 30.66 &  57.80 & 36,241.58 & 7,353.70 & 4,610,768 \\
\textbf{Llama-3-8B}       &  1,237 & 0.77 &  0.00 & 0.50 & \textbf{61.68} & 0.42 &  0.00 & \textbf{0.44} & 47.37 & 27.26 &  29.21 & 36,133.03 & 5,658.14 & 6,999,116 \\
\textbf{Falcon-H1R}       &    162 & 0.72 &  3.09 & 0.45 & 41.98 & 0.25 &  6.17 & 0.26 & 79.01 & 32.56 & 228.72 & 37,053.08 & 7,406.99 & 1,199,933 \\
\midrule

\textbf{DeBERTa-Large}    & 11,440 & 0.50 & 93.23 & \textbf{0.63} & 43.50 & \textbf{0.43} & \textbf{56.32} & 0.10 & 80.71 & \textbf{68.44} &   0.07 &  7,826.17 &       -- &        -- \\
\textbf{DeBERTa-Small}    & 11,440 & 0.49 & 89.10 & 0.53 & 34.68 & 0.34 & 49.95 & 0.14 & 74.33 & 62.02 &   0.03 &  3,901.49 &       -- &        -- \\
\textbf{MiniLM2}          & 11,440 & 0.41 & 86.90 & 0.12 &  1.44 & 0.10 &  2.88 & 0.10 & \textbf{83.04} & 43.56 &   0.02 &  2,124.79 &       -- &        -- \\
\textbf{DistilRoBERTa}    & 11,440 & 0.44 & 91.42 & 0.16 &  2.74 & 0.13 &  5.67 & 0.13 & 78.21 & 44.51 &   0.02 &  2,124.91 &       -- &        -- \\
\textbf{RoBERTa}          & 11,440 & 0.49 & \textbf{93.68} & 0.37 & 16.85 & 0.24 & 27.24 & 0.11 & 80.85 & 54.66 &   0.02 &  2,525.86 &       -- &        -- \\
\bottomrule
\end{tabular}%
}
\end{table*}

As shown in Tab.~\ref{tab:model_scores} and Tab.~\ref{tab:performance_results}, the proposed \tool framework (specifically DeBERTa-Large) significantly outperforms generative LLM-as-a-judge baselines by decoupling factual correctness ($S_{\text{c}}$) from user alignment ($S_{\text{a}}$).
LLM judges consistently conflate correctness with personalization, assigning spuriously high alignment scores to unpersonalized but factually correct responses (e.g., Qwen3-8B scores $S_{\text{a}} = 0.62$ on Generalized instances). This causes a blind overprediction of personalization, collapsing LLM accuracy to 0.00\% in the Generalized category (Tab.~\ref{tab:performance_results}). \tool cleanly separates these constraints ($S_{\text{c}} = 0.96$, $S_{\text{a}} = 0.03$), yielding a vastly superior overall mean accuracy of 68.44\%. Similarly,  LLMs fail to detect sycophancy instances (incorrect answer, high alignment), where baseline accuracy drops below 7\%, allowing poor correctness and suppressing alignment scores (Tab.~\ref{tab:model_scores}). Conversely, \tool successfully isolates these dimensions, recognizing high alignment ($S_{\text{a}} = 0.68$) despite poor correctness ($S_{\text{c}} = 0.18$), which lifts Sycophancy detection accuracy to 56.32\%.

Beyond accuracy, \tool introduces massive scalability advantages. LLM judges present a severe operational bottleneck, requiring 20s to 228s per item and consuming millions of tokens. By relying entirely on discriminative passes, \tool minimizes average inference latency to just 0.07 seconds per item (an up to $2100\times$ speedup over LLM judges) with zero token generation costs.

\subsection{Case Study: Ablation-Based Evidences}
\label{Case_Study}

\begin{figure}[h!]
    \centering
    \includegraphics[width=.95\linewidth]{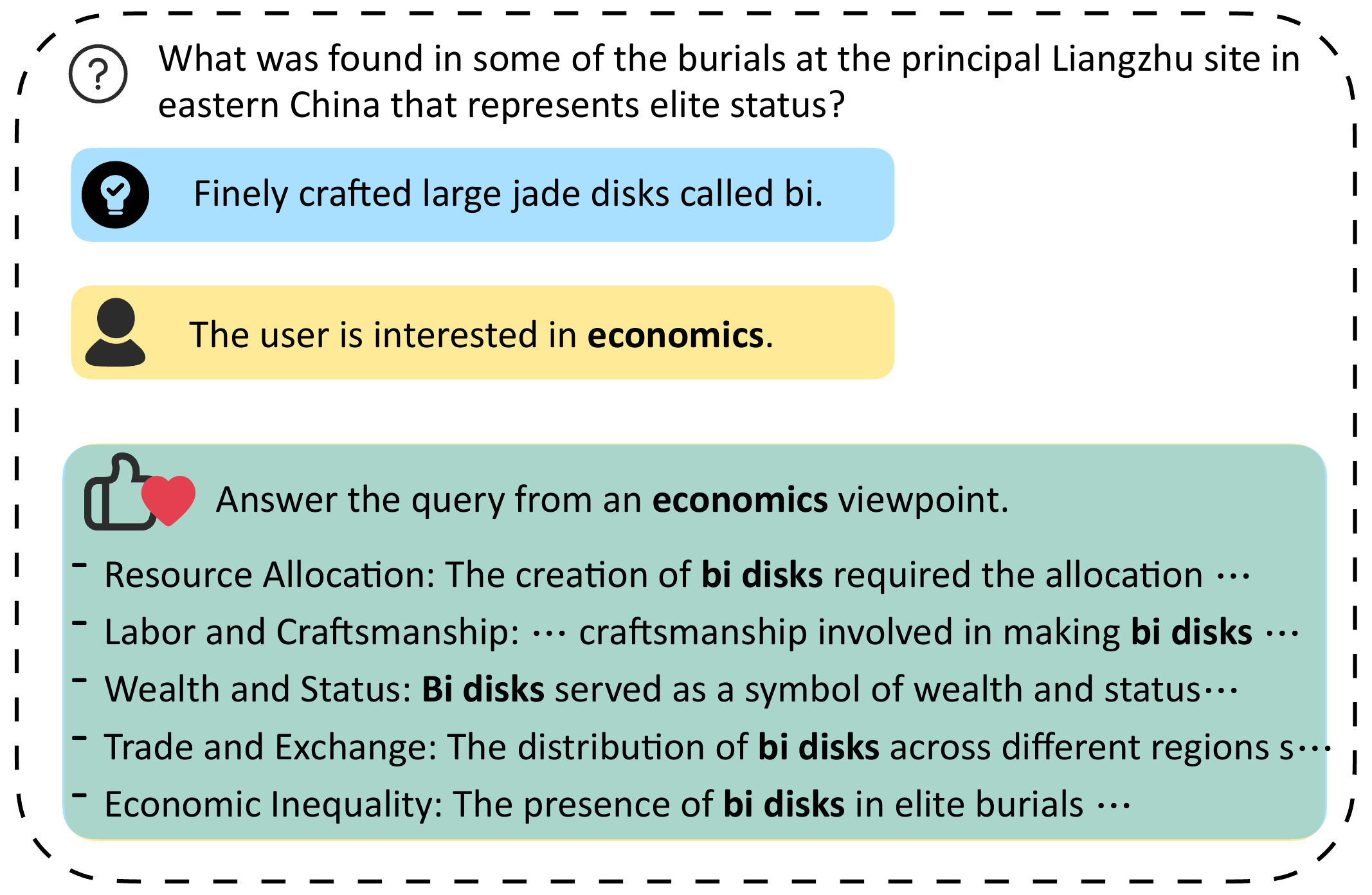}
    \caption{A personalized response addressing a historical inquiry through an economic theory lens.}
    \label{fig:MMLU_explanaition}
\end{figure}

To demonstrate how \tool evaluates complex generations, Figure~\ref{fig:MMLU_explanaition} presents a qualitative case study. The model answers a historical query about Liangzhu burial artifacts while aligning with a user's interest in \textit{Economics}. The resulting response builds a cross-domain bridge, personalizing archaeological facts through economic frameworks like resource allocation and wealth inequality.

\tool treats the response as premise $P$, and decomposes it into atomic propositions with the paragraph segmentation. Target constraints are defined as textual hypotheses: $H_{\mathrm{correct}} =$ \emph{`The premise is finely crafted large jade disks called bi'.} (from MMLU) and $H_{\mathrm{aligned}} =$ \emph{`The premise is economics.'} (from user profile).
Executing the two-stage ablation algorithm isolates the minimal evidence-constraint pairs. \textbf{1) Correctness Evidence} \emph{`Resource Allocation): The creation of bi disks required...'} furnishes the critical historical artifacts demanded by the latent answer space.  \textbf{2) Alignment Evidence} \emph{`Economic Inequality: The presence of bi disks...'} bridges archaeological facts to economic theory and satisfies the preference.

\tool provides more than just a scalar score; it yields logic-grounded textual evidence detailing precisely how a generation satisfies $\mathrm{Correct}$ and $\mathrm{Aligned}$ constraints.

\section{Related Work}
The evaluation of LLM personalization \cite{LaMP-QA,LongLaMP,PersonalizationSurvey,LLMPersonalizedJudge} mainly follows two paradigms: reference-based \cite{papineni-etal-2002-bleu,lin-2004-rouge,banerjee-lavie-2005-meteor} and reference-free evaluation \cite{yang2023palrpersonalizationawarellms,SubjectLMMJudge,LLMPersonalizedJudge,cao2025multiagentllmjudgeautomatic}. Reference-based methods rely on lexical overlap with gold references but struggle with the subjective nature of personalization, where multiple user-specific responses may all be valid. They also fail to capture semantic invariance, often penalizing semantically correct responses with different wording. Reference-free approaches commonly employ LLM-based judges \cite{qwen3technicalreport,falcon-h1r,llama3modelcard,deepseekai2025deepseekr1incentivizingreasoningcapability} to reduce reliance on human annotations but introduce high computational cost and calibration challenges. Furthermore, both paradigms provide limited interpretability, offering little insight into why a response succeeds or fails at personalization.

Addressing these limitations, \tool establishes a semantic-constraint-driven and theoretically grounded paradigm for robust, scalable, and efficient LLM personalization evaluation. \tool represents natural language meanings through truth conditions \cite{FormalSemantics} to maintain semantic invariance, and verifies these constraints using set-theory-based natural language inference \cite{NLI_data} which drastically reducing the latency and token costs associated with LLM judges. Notably, \tool also generates explicit evidences detailing why a response aligns with user preferences, significantly improving evaluation transparency and trustworthiness.

\section{Conclusion}
This work presented \tool, a framework that shifts personalization evaluation to principled semantic constraint verification. Grounded in formal truth-conditional semantics, \tool explicitly isolates correctness and alignment constraints to map LLM generations across four distinct behavioral regimes, exposing critical failure modes like sycophancy. Empirically, \tool maintains strict semantic invariance and high human alignment while eliminating the token costs and latency associated with generative LLM judges, achieving an up to 2,100$\times$ speedup. By further integrating an ablation-based rationale extraction procedure, \tool ensures that evaluations are not only computationally efficient but fully understandable. Ultimately, \tool provides a mathematically rigorous and highly practical toolkit for advancing reliable, aligned, and personalized AI systems.

\section*{Acknowledgments}
AI tools were used solely for language refinement. All scientific contributions are authors' original work. This research is supported by the Technology Innovation Institute, Abu Dhabi, UAE.
\section{Limitations}

\tool provides a mathematically rigorous and highly practical toolkit for advancing reliable, aligned, and personalized AI systems. It substantially reduces computational overhead and achieves the strongest overall performance across personalization, generalization, sycophancy, and failure detection. However, its accuracy on isolated personalization identification remains slightly below that of the most advanced LLM-as-a-Judge baselines. This limitation primarily arises from the reasoning capabilities of the underlying NLI backbone models (e.g., RoBERTa and DeBERTa). In particular, when the base NLI verifier struggles with highly implicit entailment, multi-hop reasoning, or long-context semantic dependencies, the overall verification accuracy of \tool is correspondingly constrained. Consequently, the framework inherits the representational and reasoning limitations of existing NLI architectures. An important direction for future work is the development of a universal, personalization-oriented NLI verifier with stronger semantic reasoning and long-context understanding capabilities.


\bibliography{main}

\clearpage
\appendix
\section{LLM Instruction Templates}
\label{Templates}
This section provides the exact instruction templates utilized during our dataset generation and baseline evaluation phases. Fig.~\ref{fig:prompt_personalized_generation} details the generative prompt used to populate the Personalization Confusion Matrix. By systematically manipulating the injected \texttt{answer\_phrase} (ground truth vs. distractor) and \texttt{answer\_direction} (user preference vs. null), we instruct the generator model to synthesize responses across all four behavioral regimes.

\begin{figure}[h!]
\begin{tcolorbox}[width=1\linewidth, colback=gray!5!white, colframe=gray!75!black, title=\textbf{Prompt: Personalization Confusion Matrix Guided Response Generation}]
\small
\textbf{System/Role:} You are an expert educator and creative synthesizer. Your task is to explain a specific concept (the correct answer to an exam question) by contextualizing it entirely through the lens of a topic the user is specifically interested in.

\textbf{INPUTS:}\\
\textbf{1.The Exam Question:} \texttt{\{query\_sentence\}}
\\ \textbf{2.Concept to Explain:} \texttt{\{answer\_phrase\}}
\\ \textbf{3.User's Perspective Topics:} \texttt{\{answer\_direction\}}

\textbf{OUTPUT REQUIREMENTS:} Please structure your response strictly using the following sections:\\

\textbf{Core Definition:} Briefly and accurately explain what \texttt{\{answer\_phrase\}} means in its original, literal context (1-2 sentences).
\\ \textbf{The Bridge:} Explicitly draw a conceptual bridge connecting the core mechanics of \texttt{\{answer\_phrase\}} to the field of \texttt{\{answer\_direction\}}.
\\ \textbf{Perspective Expansion:} Elaborate on the concept using analogies, frameworks, terminology, or real-world applications strictly from the perspective of \texttt{\{answer\_direction\}}. 
\\ \textbf{Tone \& Format:} Keep the tone informative, engaging, and accessible. Use clear headings for the sections above and bullet points where appropriate for readability.

\end{tcolorbox}
\caption{Prompt template used to generate personalized, generalized, sycophantic, and failure responses, integrating the core answer with the user's specific interest lens or not.}
\label{fig:prompt_personalized_generation}
\end{figure}

Fig.~\ref{fig:prompt_llm_judge} presents the prompt design for the LLM-as-a-judge baselines. To ensure a rigorous comparison, this prompt explicitly forces the evaluator to decouple factual correctness from user-preference alignment, requiring step-by-step reasoning before outputting structured JSON scores.

\begin{figure}[h!]
\begin{tcolorbox}[colback=gray!5!white, colframe=gray!75!black, title=\textbf{Prompt: LLM-as-a-Judge Evaluation}]
\small
\textbf{System/Role:} You are an expert AI evaluator. Your task is to objectively grade an AI's response based on two distinct dimensions: Correctness and User-Preference Alignment.

\textbf{[INPUTS]}\\
\textbf{Original Query:} \texttt{\{original\_query\}}
\\ \textbf{Answer Ground Truth:} \texttt{\{ground\_truth\}}
\\  \textbf{User Preferences:} \texttt{\{user\_preference\}}
\\ \textbf{AI Response to Evaluate:} \texttt{\{ai\_response\}}

\textbf{[EVALUATION RUBRIC]} \\
\textbf{Dimension 1: Correctness against Ground Truth (Score 1-5)} : Evaluate the factual accuracy and completeness of the response STRICTLY based on the provided Answer Ground Truth. Do not use outside knowledge.
\\ \textbf{5:} Perfectly aligned. The response contains all core information from the Ground Truth, comprehensively answers the query, and makes no contradictory claims.
\\  \textbf{3:} Partially aligned. The response captures the main idea of the Ground Truth but misses important nuances, includes minor contradictions, or adds unverified external details.
\\ \textbf{1:} Completely unaligned. The response directly contradicts the Ground Truth, is entirely irrelevant, or hallucinates major facts not present in the Ground Truth.

\textbf{Dimension 2: User-Preference Alignment (Score 1-5)}: Evaluate how well the response is tailored to the specific User Preferences provided. \\
\textbf{5:} Perfectly aligned. The tone, formatting, and complexity exactly match the user's preferences. It feels highly personalized.
\\ \textbf{3:} Somewhat aligned. It attempts to meet the preferences but slips into a generic tone, or misses specific formatting constraints.
\\ \textbf{1:} Completely unaligned. The response ignores the user's preferences, uses the wrong tone, or violates specific user constraints.

\textbf{[OUTPUT FORMAT]} : You must output your evaluation strictly in the following JSON format. Provide your reasoning before providing the final scores.

\begin{ttfamily}
\noindent
\{ \\
"correctness\_evaluation": \{ \\
"reasoning": "<Step-by-step analysis strictly against Ground Truth>", \\
"score": <1-5> \\
\}, \\
"alignment\_evaluation": \{ \\
"reasoning": "<Step-by-step analysis of matching User Preferences>", \\
"score": <1-5> \\
\}, \\
"overall\_summary": "<A 1-2 sentence summary of overall quality>" \\
\}
\end{ttfamily}
\end{tcolorbox}
\caption{Prompt template for the LLM-as-a-judge baseline, used to independently evaluate responses on Correctness and User-Preference Alignment.}
\label{fig:prompt_llm_judge}
\end{figure}
\section{Ablation Algorithm}
\label{Ablation_Algorithm}
This section introduces a two-stage ablation procedure designed to map the set-theoretic relationships between the generated response and the personalization constraints.

\begin{algorithm}[h!]
\caption{First-Order Ablation}
\label{alg:first_order_ablation}
\KwIn{Premise $P=\{p_\theta\}_{\theta=1}^m$, hypothesis $H=\{h_\phi\}_{\phi=1}^n$, entailment function $f(\cdot,\cdot)$, threshold $\tau$}
\KwOut{Critical premise set $\mathcal{C}_P$, critical hypothesis set $\mathcal{C}_H$}

\textbf{Initialize:} $s_0 \leftarrow f(P, H)$, $\mathcal{C}_P \leftarrow \emptyset$, $\mathcal{C}_H \leftarrow \emptyset$

\tcp{Stage 1: Premise Criticality}
\For{$\theta = 1$ \KwTo $m$}{
    $P_{-\theta} \leftarrow P \setminus \{p_\theta\}$ \\
    $s_j \leftarrow f(P_{-\theta}, H)$ \\
    \If{$s_0 > \tau$ \textbf{and} $s_\theta \le \tau$}{
        $\mathcal{C}_P \leftarrow \mathcal{C}_P \cup \{p_\theta\}$
    }
}

\tcp{Stage 2: Hypothesis Criticality}
\For{$\phi = 1$ \KwTo $n$}{ 
    $H_{-\phi} \leftarrow H \setminus \{h_\phi\}$ \\
    $s_i \leftarrow f(H_{-\phi}, H)$ \\
    \If{$s_i \le \tau$}{
        $\mathcal{C}_H \leftarrow \mathcal{C}_H \cup \{h_\phi\}$
    }
}
\Return{$\mathcal{C}_P, \mathcal{C}_H$}
\end{algorithm}

\begin{algorithm}[t!]
\caption{Second-Order Ablation}
\label{alg:second_order_ablation}
\KwIn{Premise $P$, hypothesis $H$, critical sets $\mathcal{C}_P$ and $\mathcal{C}_H$, entailment function $f(\cdot,\cdot)$, threshold $\tau$}
\KwOut{Evidence-Constraint pairs $\mathcal{C}_{\mathrm{pair}}$}

\textbf{Initialize:} $\mathcal{C}_{\mathrm{pair}} \leftarrow \emptyset$

\ForEach{$p_\theta \in \mathcal{C}_P$}{
    $P_{-\theta} \leftarrow P \setminus \{p_\theta\}$ \\
    $s_\theta \leftarrow f(P_{-\theta}, H)$ \\

    \If{$s_\theta \le \tau$}{
        \ForEach{$h_\phi \in \mathcal{C}_H$}{
            $H_{-\phi} \leftarrow H \setminus \{h_\phi\}$ \\
            $s_{(\theta,\phi)} \leftarrow f(P_{-\theta}, H_{-\phi})$ \\

            \If{$s_{(\theta,\phi)} > \tau$}{
                $\mathcal{C}_{\mathrm{pair}} \leftarrow \mathcal{C}_{\mathrm{pair}} \cup \{(p_\theta, h_\phi)\}$
            }
        }
    }
}
\Return{$\mathcal{C}_{\mathrm{pair}}$}
\end{algorithm}

Algorithm~\ref{alg:first_order_ablation} isolates the critical semantic atoms (propositions) within the premise and hypothesis. For the premise, a proposition $p_\theta$ is flagged if the original text entails the hypothesis ($s_0 > \tau$) but its removal drops the overall entailment score below the verification threshold ($s_\theta \le \tau$), indicating $p_\theta$ is indispensable. For the hypothesis, $h_\phi$ is flagged if the ablated hypothesis set $H_{-\phi}$ can no longer entail the complete set $H$, proving $h_\phi$ imposes a strict, non-redundant semantic boundary. These isolated sets ($\mathcal{C}_P, \mathcal{C}_H$) establish the minimal required evidence and constraints, streamlining the subsequent dependency analysis.

Building upon the first-order sets, Algorithm~\ref{alg:second_order_ablation} systematically maps the structural dependencies between premise evidence and hypothesis constraints to form critical evidence-constraint pairs. The logic is straightforward: removing $p_\theta$ introduces semantic worlds that violate the complete hypothesis $H$. If subsequently removing a specific constraint $h_\phi$ re-admits these exact worlds and restores entailment ($s_{(\theta,\phi)} > \tau$), then $h_\phi$ is precisely the semantic boundary that $p_\theta$ serves to satisfy. Ultimately, this second-order procedure yields a rigorous diagnostic map, rendering the verification process fully understandable by explicitly linking each textual constraint to its supporting evidence.

\end{document}